\documentclass[sigchi-a]{acmart}

\def\BibTeX{{\rm B\kern-.05em{\sc i\kern-.025em b}\kern-.08emT\kern-.1667em\lower.7ex\hbox{E}\kern-.125emX}}

\copyrightyear{2019}
\acmYear{2019}
\setcopyright{acmlicensed}
\acmConference[HCML Perspectives Workshop]{HCML Perspectives Workshop at CHI 2019}{May 04, 2019}{Glasgow}

\usepackage{xspace}
\definecolor{col1}{rgb}  {0.9, 0.5, 0.1}

\definecolor{col2}{rgb}  {0.1, 0.5, 1.0}

\definecolor{awesome}{rgb}{1.0, 0.13, 0.32}

\newcommand\system{PreCall\xspace}

\begin{document}

%
\title{\system: A Visual Interface for Threshold Optimization in ML Model Selection}

%
\author{Christoph Kinkeldey}
\email{christoph.kinkeldey@fu-berlin.de}
\affiliation{%
  \institution{Human-Centered Computing\\Freie Universit\"at Berlin}
  \streetaddress{K\"onigin-Luise-Straße 24-26}
  \city{Berlin}
  \state{Germany}
  \postcode{14195}
}

\author{Claudia M\"uller-Birn}
\email{clmb@inf.fu-berlin.de}
\affiliation{%
  \institution{Human-Centered Computing\\Freie Universit\"at Berlin}
  \streetaddress{K\"onigin-Luise-Straße 24-26}
  \city{Berlin}
  \state{Germany}
  \postcode{14195}
}

\author{Tom G\"ulenman}
\email{tom.guelenman@fu-berlin.de}
\affiliation{%
  \institution{Human-Centered Computing\\Freie Universit\"at Berlin}
  \streetaddress{K\"onigin-Luise-Straße 24-26}
  \city{Berlin}
  \state{Germany}
  \postcode{14195}
}

\author{Jesse Josua Benjamin}
\email{jesse.benjamin@fu-berlin.de}
\affiliation{%
  \institution{Human-Centered Computing\\Freie Universit\"at Berlin}
  \streetaddress{K\"onigin-Luise-Straße 24-26}
  \city{Berlin}
  \state{Germany}
  \postcode{14195}
}

\author{Aaron Halfaker}
\email{ahalfaker@wikimedia.org}
\affiliation{%
  \institution{Wikimedia Foundation}
  \streetaddress{1 Montgomery Street: Suite 1600}
  \city{San Francisco, CA}
  \state{USA}
  \postcode{94104}
}

%
\renewcommand{\shortauthors}{Kinkeldey, et al.}


\begin{abstract}
Machine learning systems are ubiquitous in various kinds of digital applications and have a huge impact on our everyday life. But a lack of explainability and interpretability of such systems hinders meaningful participation by people, especially by those without a technical background. Interactive visual interfaces (e.g., providing means for manipulating parameters in the user interface) can help tackle this challenge. In this paper we present \system, an interactive visual interface for ORES, a machine learning-based web service for Wikimedia projects such as Wikipedia. While ORES can be used for a number of settings, it can be challenging to translate requirements from the application domain into formal parameter sets needed to configure the ORES models. Assisting Wikipedia editors in finding damaging edits, for example, can be realized at various stages of automatization, which might impact the precision of the applied model. Our prototype \system attempts to close this translation gap by interactively visualizing the relationship between major model metrics (recall, precision, false positive rate) and a parameter (the threshold between valuable and damaging edits). Furthermore, \system visualizes the probable results for the current model configuration to improve the human's understanding of the relationship between metrics and outcome when using ORES. We describe \system's components and present a use case that highlights the benefits of our approach. Finally, we pose further research questions we would like to discuss during the workshop. 
\end{abstract}

%
%



%

%

%
\maketitle

\section{Introduction}

\begin{marginfigure}
  \includegraphics[width=0.9\linewidth]{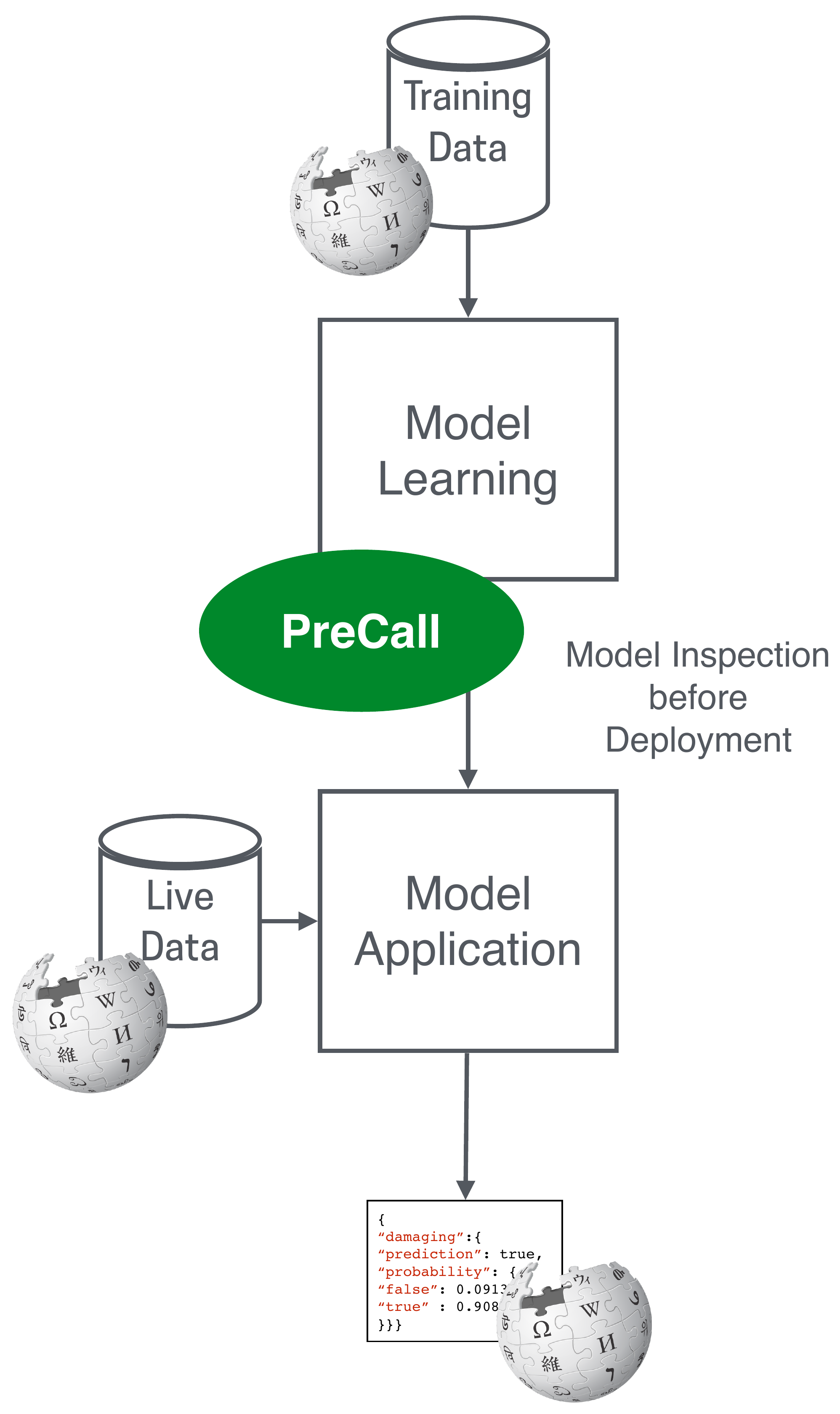}
  \Description{ORES}
  \caption{The \system approach can be localized at the deployment stage of a machine learning model.}
  \label{fig:ores}
\end{marginfigure}

With rising concerns over the utilization of machine learning (ML) algorithms in everyday activities as well as high-stakes environments such as the law~\cite{angwin_machine_2016}, new directions for the development and deployment of algorithmic systems for non-technical groups have emerged. On the one hand, legislative imperatives such as the GDPR's implicit right to explanation~\cite{selbst_meaningful_2018} have led to design strategies that supply explanations for ML algorithms in user interfaces. However, these are predominantly created by technical experts and can prove unsuitable for non-technical groups ~\cite{miller_explainable_2017,edwards_slave_2017}. As an extension to explanations, interactive approaches, such as sample review, feedback assignment, model inspection, and task overview, have been suggested in order to foster more meaningful participation in systems that use ML algorithms ~\cite{dudley_review_2018}. We see these interactive approaches as integral to the goal of making ML more accessible. In our paper, we focus on the model inspection facet of machine learning systems (cp.~\autoref{fig:ores}). We present an interactive user interface for the ML back-end service ORES that supports developers in selecting a model configuration meeting their requirements.

\section{Use Case: ORES}
\begin{marginfigure}
  \centering
  \includegraphics[width=\linewidth]{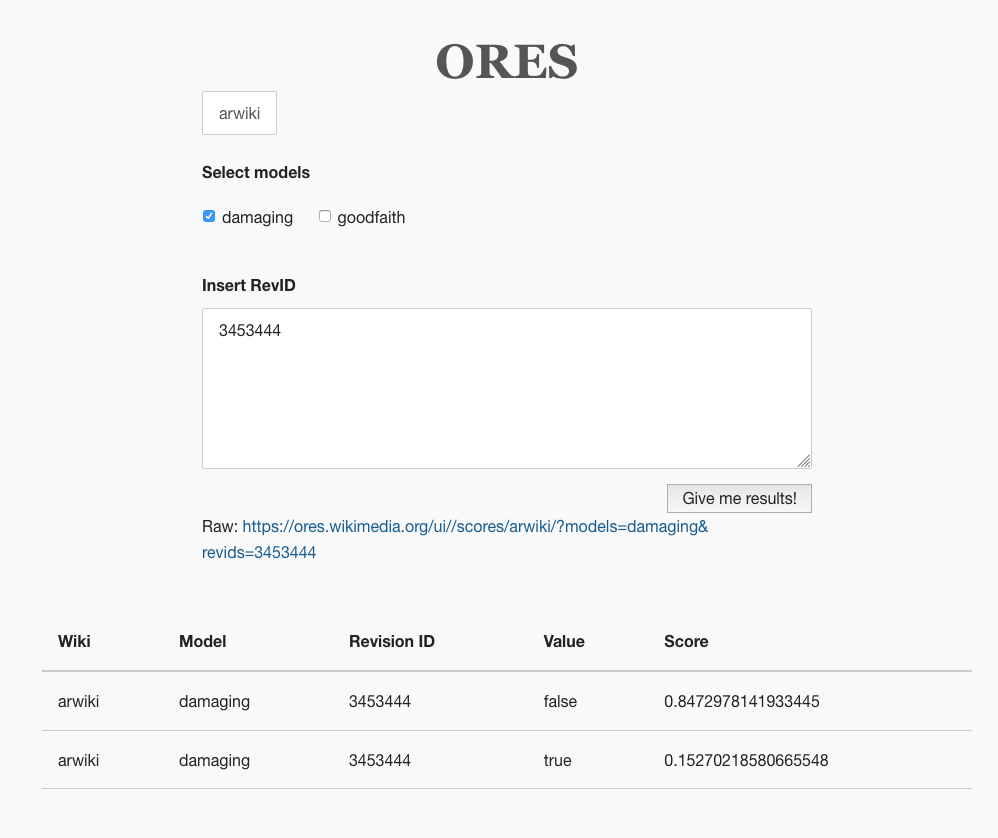}
  \caption{Current interface for the ORES damaging detection system, (\url{https://ores.wikimedia.org/ui/}).}
  \Description{Description.}
  \label{fig:ores-interface}
\end{marginfigure}

Only a few years after its inception, the number of active volunteers in Wikipedia grew exponentially. At the same time, this success lead to increasing vandalism in Wikipedia. The English Wikipedia, for example, receives over 150 thousand new edits every day, which go live immediately and without verification. Wikipedians accept this risk of an open encyclopedia but work tirelessly to maintain quality. However, it has become no longer possible to do so manually. Due to its ongoing growth, Wikipedia entered into a phase of automation, and many quality control tools, such as ClueBot NG\footnote{ClueBot NG pre-classifies edits in Wikipedia by with a Bayesian Classifiers to reduce the percentage of false positives. Then an artificial neural network is used to classify the detected vandalism. ClueBot NG generates a vandalism probability for each edit.} were developed. However, developers in Wikipedia are volunteers who often learn Javascript/Python by themselves. They are non-technical experts and do not have deep enough technical expertise in machine learning terms and practices, and therefore, they lack the expertise to develop the machine prediction models necessary to power quality control tools. The Wikimedia Scoring Platform team~\cite{Wikimedia} tackled this challenge and developed ORES, a machine prediction service developed and maintained by professionals, but intended to be used by volunteer tool developers.
ORES is a back-end service that allows other tools to simply provide one or more revision ID(s) and receive the probability scores for the respective revision(s) as being ``damaging'' or ''not damaging''. As of now, ORES offers a web API to make use of its models, and a very simple user interface exists which allows people to retrieve scoring information about edits across a multitude of wikis (\autoref{fig:ores-interface}).

Developers who want to apply the ORES damaging prediction need to choose a threshold of confidence that supports the work practices they are designing for.  
But 
inspecting the model and determining an appropriate choice for a specific purpose is not well supported for non-ML experienced developers. In the next section, we describe existing challenges that occur when employing ORES as quality control system.

\section{Human-Centered Optimization of Model Configuration}
Halfaker et al.~\cite{halfaker2018ores} describe the case of PatruBot from Spanish Wikipedia. An editor developed PatruBot based on ORES to revert damaging edits in Spanish Wikipedia automatically. However, soon after its initiation, the Wikimedia Scoring Platform team received complaints from editors who did not understand why PatruBot reverted their edits. After investigation, it showed that the bot reverted edits that passed a low threshold likelihood of being damaging. In case of a fully automated quality control process, the model needs to be optimized to a high \emph{precision}, i.e., only damaging edits are flagged, which results in a lower \emph{recall}, i.e., some suspicious edits remain undetected. What we derive from this case is that even with knowledge about ORES, it is not straightforward for people to come up with a confidence threshold that meets their operational requirements (e.g., high precision at the cost of recall).  The interplay of model fitness metrics and expectations requires interpretation on a case-by-case basis. 

We see ORES as a particularly fruitful setting for developing interfaces that lower the barrier for non-technical community access to ML. Accordingly, we were motivated to prototype \system, an interactive visual interface to support non-technical experts in developing a mental model of the ORES classifier when selecting a suitable model configuration for their application.
Previous research has shown that interactive visualizations enabling people to tweak ML systems help them to make more effective use of ML-services~\cite{Kapoor:2010in, Amershi:2015cu}. In our research, we build upon this line of research and seek to support people in finding optimal model configurations for ORES that meet their requirements, without having to understand how exactly the system works internally. 

\section{The \system Visual Interface Design}
\begin{marginfigure}
  \centering
  \includegraphics[width=\linewidth]{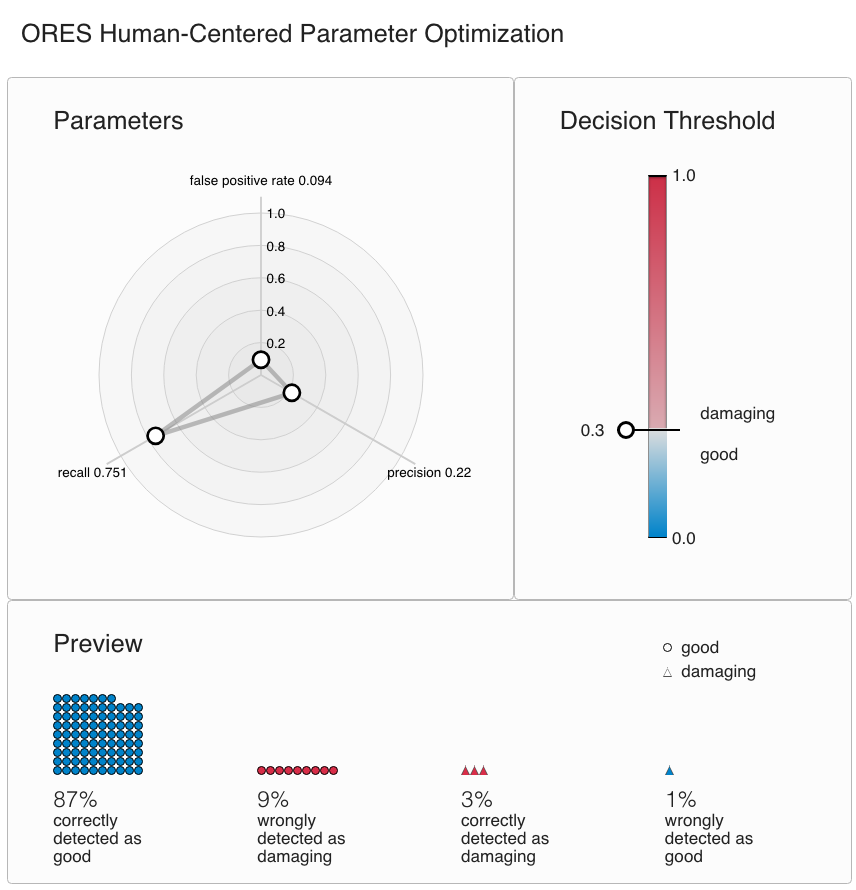}
  \caption{Interactive user interface of \system consisting of a parameter view (top) including a decision threshold slider, and a preview of the results (bottom).}
  \Description{Description.}
  \label{fig:precall-gui}
\end{marginfigure}

The visual interface aims to support the interpretation of different configurations of the damaging classifier expressed by model fitness metrics, and the confidence threshold that defines which score separates good from damaging edits. We designed two views covering the main tasks: a parameter view to inform a person about possible configurations of the damaging model, and a preview of the expected outcome of the classifier (\autoref{fig:precall-gui}).

\subsection{Parameter View}
The first aim of the visual interface was to show the relationship of the three major fitness metrics of the ORES damaging model: recall, precision, and false-positive rate. In the GUI they are represented as three axes of a radar chart (\autoref{fig:precall-gui}, top left). A person can vary any metric and the other two are updated instantly. The second aim was to demonstrate how the confidence threshold relates to the model metrics. A slider next to the radar chart represents the threshold which determines if an edit is declared as good or damaging (\autoref{fig:precall-gui}, top right). A color gradient illustrates the fact that the transition from good to damaging edits is fluid, i.e., there is a range of uncertainty. Changing the threshold in the slider also immediately changes the values in the radar chart. This way, interaction facilitates the exploration of different thresholds and model metrics, as well as their interdependence.

\subsection{Preview of Results}
Another crucial goal of \system is to demonstrate how the outcome of the model changes with different configurations. The view on the bottom (\autoref{fig:precall-gui}, bottom) shows the predicted outcome for the chosen configuration as stacked symbols. This view is designed to provide an intuitive representation of the expected result to let the user quickly grasp the number of elements belonging to the different groups: true negative, false positive, true positive, and false negative flags of edits. Color expresses how the algorithm tagged the edits: good (blue) and damaging (red). The shape of the elements represents their true state: good  (circle) and actually damaging (triangle) edits. Compared to the common way of showing classification results in a confusion matrix (e.g.~\cite{Kapoor:2010in}), we hypothesize that this visualization provides a more intuitive representation of a classification outcome. Moreover, by also adapting instantly, this preview further strengthens \system's interpretive support by describing the relationship between model configuration and expected output.

\subsection{Determining a Suitable Model Configuration for Semi-Automated Edit Review}   

\begin{marginfigure}
  \centering
  \includegraphics[width=\linewidth]{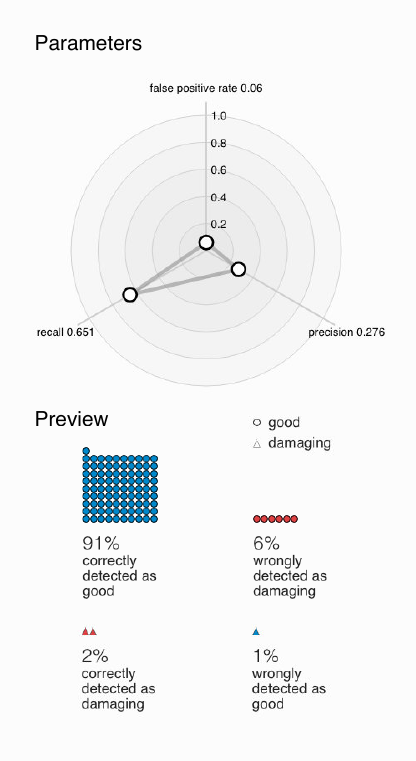}
  \caption{The result for the use case of semi-automated edit review: a threshold of 0.4 minimizes the number of detected damaging edits.}
  \Description{Description.}
  \label{fig:usecase}
\end{marginfigure}

Based on the use case described above, the semi-automated review of edits, we demonstrate of how \system can help finding the optimal model configuration for a specific application:

\begin{enumerate}
\item We start with a threshold of 0.5, which results in recall of $0.569$, precision of $0.347$, and false positive rate of $0.038$. With this threshold, the number of falsely detected good edits is still quite high (''2\% wrongly detected as good''), as we have the same amount of correctly detected damaging edits.
\item In order to let the system find further damaging edits, we decrease the decision threshold to 0.3. The parameter view reveals that recall goes up ($0.751$) and precision down ($0.22$). The fraction of ``wrongly detected as good'' edits went down to 1\%, however there are still 12\% of edits altogether that are (correctly and falsely) detected as damaging and would have to be reviewed manually. 
\item Trying out other thresholds, we find a better choice: with a threshold of $0.4$ the number of edits that are detected as damaging is minimized to 8\% (with 6\% wrongly and 2\% correctly detected, see \autoref{fig:usecase}). Given the 91\% of edits correctly detected as good this is a better outcome for our purpose of reviewing a small number of uncertain edits among a large set of edits.
\item After gaining a better understanding of the model characteristics, we are satisfied with this payoff and decide to use the chosen configuration to check new data for damaging edits.
\end{enumerate}

This scenario shows how \system, with its integrated visual approach, is intended to support the configuration of the ORES damaging model. We hope we can show in planned user studies that \system helps people build a meaningful understanding of model metrics and confidence threshold, their relationship, and how they affect the possible outcome.

\section{Discussion}
In this paper we described the context, the requirements, and the current design rationales of the work-in-progress development of \system. The main goal of the approach is to support the editors in Wikimedia projects, i.e. non-technical experts, in arriving at a case-specific meaningful interpretation when selecting a configuration of the ORES damaging model that fits their requirements. The current prototype serves as a demonstration of the concept and as testing platform for the wider community. 

For evaluation of \system's potential to support the interpretation of the ML model for specific case-by-case usage of ORES, we envision a qualitative user study with Wikipedia editors. A particular concern is the level of abstraction \system should provide, such as whether our inclusion of measures like precision and recall is interpretable for Wikipedia tool developers. Therefore, our study should also compare ours to more abstract approaches such as an interactive confusion matrix as proposed by Kapoor et al.~\cite{Kapoor:2010in}. Another possible qualitative dimension to our studies is comparing the understanding gained by using \system as opposed to reading the officially supplied documentation for ORES parameters (e.g., \url{https://www.mediawiki.org/wiki/ORES/Thresholds}). 
If our approach turns out to be useful, a future goal would be to provide the Wikimedia community with an enhanced version of \system for long-term field studies. In this way, we hope to improve our understanding of how such visual interfaces can impact the acceptance and usage rate of ML-systems in the community.

We see visual parameter selection support approaches like \system as valuable contributions to participatory use of machine learning systems. In this workshop we would like to discuss our strategy of facilitating better interpretability of machine learning systems, without necessarily pursuing the goal of making them entirely transparent. We are convinced that following this strategy, visual approaches have the potential to foster a better understanding of machine learning-based decision making. 

%
\bibliographystyle{ACM-Reference-Format}
\bibliography{precall}

%
\appendix

\end{document}